\documentclass{sig-alternate-05-2015}
\usepackage{icvgip}
\usepackage{graphicx}
\usepackage{multirow}

\usepackage{amsmath,amssymb} 
\usepackage{color}
\usepackage{graphicx}
\usepackage{amsmath}
\usepackage{amssymb}
\usepackage{array}
\usepackage{algorithm}
\usepackage{algorithmic}
\usepackage{xspace}
\usepackage[abs]{overpic}

\def\ie{ie.\@\xspace}

\begin{document}
 
\icvgipfinalcopy
 
 


%
 
\def\icvgipPaperID{326}

\title{Deep fusion of visual signatures\\ for client-server facial analysis}  

\numberofauthors{3} %
\author{
\alignauthor
Binod Bhattarai\titlenote{}\\
       \affaddr{Normandie Univ, UNICAEN, ENSICAEN, CNRS, GREYC}\\
              \email{binod.bhattarai@unicaen.fr}
\alignauthor
Gaurav Sharma\titlenote{}\\
       \affaddr{Computer Sc. \& Engg.}\\
       \affaddr{IIT Kanpur, India}\\
       \email{grv@cse.iitk.ac.in}
\alignauthor Frederic Jurie\titlenote{}\\
        \affaddr{Normandie Univ, UNICAEN, ENSICAEN, CNRS, GREYC}\\
        \email{frederic.jurie@unicaen.fr}
 }
 
\maketitle
\def\etal{et al\onedot}
\def\etc{etc\onedot}
\def\ie{i.e\onedot}
\def\eg{e.g\onedot}
\def\cf{cf\onedot}
\def\vs{vs\onedot}
\def\grad{\nabla}
\def\b{\textbf{b}}
\def\v{\textbf{v}}
\def\a{\boldsymbol{\alpha}}
\def\sign{\textrm{sign}}
\def\pd{\partial}
\def\T{\mathcal{T}}
\def\R{\mathbb{R}}
\def\Reg{\mathcal{R}}
\def\X{\mathcal{X}}
\def\I{\mathcal{I}}
\def\F{\mathcal{F}}
\def\Obj{\mathcal{O}}
\def\V{\mathcal{V}}
\def\Lz{L_0}
\def\Lt{L_t}
\def\Dz{{D_0}}
\def\Do{{D_1}}
\def\Dt{{D_2}}
\def\w{\textbf{w}}
\def\c{\textbf{c}}
\def\1{\textbf{1}}
\def\x{\textbf{x}}
\def\c{\textbf{c}}
\def\s{\textbf{s}}
\def\d{\boldsymbol{\delta}}
\def\y{\textbf{y}}
\def\l{\textbf{l}}
\def\ock{$1$-call@$K$}
\def\oc{$1$-call@}
\def\TODO{\textcolor{red}{TODO} }

Facial analysis is a key technology for enabling human-machine interaction. In this context, we present a client-server framework, where a client transmits the signature of a face to be analyzed to the server, and, in return, the server sends back various information describing the face e.g.\ is the person male or female, is she/he bald, does he have a mustache, etc. We assume that a client can compute one (or a combination) of visual features; from very simple and efficient features, like Local Binary Patterns, to more complex and  computationally heavy, like Fisher Vectors and CNN based, depending on the computing resources available. The challenge addressed in this paper is to design a common universal representation such that a single merged signature is transmitted to the server, whatever be the type and number of features computed by the client, ensuring nonetheless an optimal performance. Our solution is based on learning of a common optimal subspace for aligning the different face features and merging them into a universal signature. We have validated the proposed method on the challenging CelebA dataset, on which our method outperforms existing state-of-art methods when rich representation is available at test time, while giving  competitive performance when only simple signatures (like LBP) are available at test time due to resource constraints on the client.

\section{Introduction}
\label{sec:intro}
We propose a novel method in a heterogeneous server-client framework for the challenging and important task of analyzing images of faces. Facial analysis is a key ingredient for assistive computer vision and human-machine interaction methods, and systems and incorporating high-performing methods in daily life devices is a challenging task. The objective of the present paper is to develop state-of-the-art technologies for recognizing facial expressions and facial attributes on mobile and low cost devices. Depending on their computing resources, the clients (i.e.\ the devices on which the face image is taken) are capable of computing different types of face signatures, from the simplest ones (e.g.\ LPB) to the most complex ones (e.g.\ very deep CNN features), and should be able to eventually combine them into a single rich signature. Moreover, it is convenient if the face analyzer, which might require significant computing resources, is implemented on a server receiving face signatures and computing facial expressions and attributes from these signatures. Keeping the computation of the signatures on the client is safer in terms of privacy, as the original images are not transmitted, and keeping the analysis part on the server is also beneficial for easy model upgrades in the future. To limit the transmission costs, the signatures have to be made as compact as possible. In summary, the technology needed for this scenario has to be able to merge the different available features -- the number of features available at test time is not known in advance but is dependent on the computing resources available on the client -- producing a unique rich and compact signature of the face, which can be transmitted and analyzed by a server. Ideally, we would like the  universal signature to have the following properties: when all the features are available, we would like the performance of the signature to be better than the one of a system specifically optimized for any single type of feature. In addition, we would like to have reasonable performance when only one type of feature is available at test time. 

For developing such a system,  we propose a {\em hybrid deep neural network} and give a method to carefully fine-tune the network parameters while learning with all or a subset of features available. Thus, the proposed network can process a number of wide ranges of feature types such as hand-crafted LBP and FV, or even CNN features which are learned end-to-end.

While CNNs have been quite successful in  computer vision~\cite{krizhevsky2012imagenet}, representing images with CNN features is relatively time consuming, much more than some simple hand-crafted features such as LBP. Thus, the use of CNN in real-time applications is still not feasible. In addition, the use of robust hand-crafted features such as FV in hybrid architectures can give performance comparable to Deep CNN features~\cite{perronnin2015fisher}. The main advantage of learning hybrid architectures is to avoid having large numbers of convolutional and pooling layers. Again from~\cite{perronnin2015fisher}, we can also observe that hybrid architectures improve the performance of hand-crafted features e.g.\  FVs. Therefore, hybrid architectures are useful for the cases where only hand-crafted features, and not the original images, are available during training and testing time. This scenario is useful when it is not possible to share training images due to copyright or privacy issues. 

Hybrid networks are particularly adapted to our client-server setting. The client may send image descriptors either in the form of some hand-crafted features or CNN features or all of them, depending on the available computing power. The server has to make correct predictions with any number and combination of features from the client. The naive solution would be to train classification model for the type of  features as well as for any of their combinations and place them in the server. This will increase the number of model parameters exponentially with the number of different feature types. The proposed hybrid network aligns the different feature before fusing them in a unique signature.

\begin{figure*}[t]
\centering
\includegraphics[width=0.8\textwidth, trim=0 10 0 0, clip]{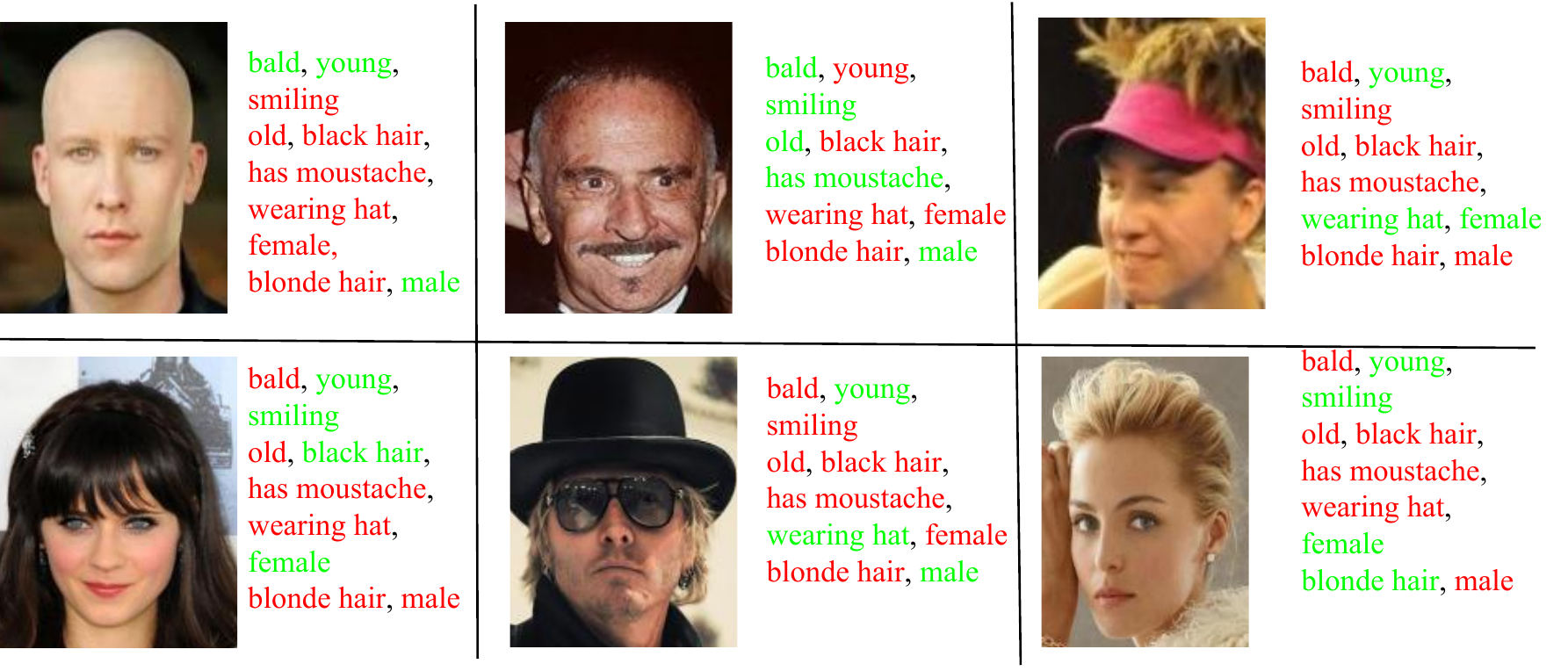} 
\vspace{-0.4em}
\caption{
Randomly sampled images of CelebA and a subset of attributes. Green color attributes are relevant for the image whereas red color attributes are irrelevant (better viewed in color).
}
\vspace{-1em}
\label{fig:illus}
\end{figure*}

The main contribution of the paper is a novel multi-features fusion hybrid deep network, which can accept a number of wide ranges of feature types and fuse them in 
an optimal way. The proposed network first processes 
the different features with feature specific layers which are then followed by layers shared by all feature types. The former layer(s) generate(s) compact and discriminative 
signatures while the later ones process the signatures to make predictions for the faces. 
We learn both feature specific parameters and shared parameters to minimize the loss function using back propagation in such a way that all the component features 
are aligned in a shared discriminative subspace.
During test time, even if all the features are not available, e.g.\ due to computation limitations, the network can make good predictions with graceful degradation depending 
on the number of features missing. 

The thorough experimental validation provided, demonstrates that the proposed architecture gives state-of-the art result on attributes prediction on the CelabA dataset when all the features are available. The method also performs competitively when the number of features available is less i.e.\ in a resource-constrained situation.

The rest of the paper is organized	 as follows: Sec. 2 presents the related works, Sec. 3 gives the details of our approach while Sec. 4 presents the experimental validation.

\section{Related Works}
\label{sec:related_works}
In this section we review some of the works which are, on one side,  related to hybrid architectures or, on the other side, related to multimodal fusion and face attribute classification. Apart from face attributes classification, other critical applications on faces are: large scale face retrieval~\cite{bhattarai2014some,bhattarai2016cvpr}, face verification~\cite{lhs_cviu_2016,simonyan2013fisher,parkhi2015deep,taigman2014deepface},   age estimation~\cite{bhattarai2016joint,guo2014study}, etc. For more details on the application of faces and comprehensive comparison of recent works, we suggest the readers refer  ~\cite{learned2016labeled}.
\vspace{0.8em} \\
\textbf{Hybrid Architectures.}
One of the closest works to our work is from Perronnin et al.~\cite{perronnin2015fisher}. The main idea behind their work is 
to use Fisher Vectors as input to  Neural Networks (NN) having few fully connected (supervised) layers (up to 3) and to learn the parameters of these layers to minimize the loss function. The parameters are optimized using back propagation.  
Unlike their architecture, our network takes a number of wide range of hand-crafted features including FVs, but not only. In addition, our architecture is also equipped with both feature specific parameters and common parameters. We have designed our network in such a way that the input 
features are aligned to each other in their sub-spaces. The advantage of such alignments is that our system can give good performance even when 
a single type of feature is present at test time. Moreover, such ability makes our system feature independent i.e. it can properly handle 
any types of features it encounters.

There are some works, such as~\cite{uricchio2015fisher}, which, instead of taking hand-crafted features as input, takes CNN features and compute FVs in the context of efficient image retrieval and image tagging. This approach improves the performance of CNNs and attains state-of-art performance, showing that
not only FVs but also CNNs benefit from hybrid architecture.
\vspace{0.8em} \\
\textbf{Face Attribute Classification.}
Some of the earliest and seminal work on facial attribute classification is the  works from Kumar et al.~\cite{kumar2008facetracer,kumar2009attribute}. 
Both of their papers use hand-crafted low-level features to represent faces, sampled with AdaBoost in order to discover the most discriminative ones for a given attribute, 
and train binary SVM classifiers on this subset of features to perform attribute classification. 
The current state-of-art method of Liu~et al.~\cite{liu2015deep} uses two deep networks, one for face localization and another for identity based 
face classification. The penultimate layer of the identity classification network is taken as the face representation,  and a binary  SVM classifier is trained to 
perform an attribute classification. Some other recent state-of-the-art methods such as PANDA~\cite{zhang2014panda}, Gated ConvNet~\cite{kang2015face}, etc.  also use deep learning
to learn the image representation and do attribute classifications on it. From these works, we can observe that either hand-crafted features  or CNN features are used for attribute
classification. From our knowledge, the proposed method is the first to learn a hybrid structure combining multiple hand-crafted and CNN 
features for facial attribute classification. Moreover, most of the mentioned works here are performing binary attribute classification while we are predicting multiple attributes of faces.
\vspace{0.8em} \\
\textbf{Multi-modal fusion.} Recently Neverova~ et al.~\cite{neverova2015oddrop} proposed a method called {\em Mod-Drop} to fuse information from multiple sources. Their main idea is to take a batch of examples from one source at a time and feed into the network to learn the parameters,  instead of taking examples from all the sources. The main drawbacks of their approach is, when a new source is encountered and is to be fused, it requires to re-trainthe whole network. Some other recent works such as ~\cite{kahou2013combining,srivastava2012multimodal,wu2014exploring,ngiam2011multimodal} fuse multiple sources of information to improve the performance of the final result. None of these works evaluated the performance of component sources or their possible combinations after fusion.

\section{Approach}
\label{sec:approach}
As mentioned before, a key challenge addressed in this paper is to learn an optimal way to fuse several image features into a common signature, through the use of a hybrid fully connected deep network. This section presents the proposed method in detail, explains how to learn the parameters and gives technical details regarding the architecture.

\subsection{Network architecture}

Fig.~\ref{fig:illus} shows a schematic diagram of the proposed network. A, B and C denote the different feature types to be aligned and fused, which are the input to the network. 
We recall that all or only a subset of the features can be available depending on the computing resources of the client. While we show a network with 3 features types, more can be used with similar layers for the new features. The key idea here is to train a single network which consists of feature specific layers (shown in blue), to be implanted on the clients, and common layers (shown in black), to be implanted on the server. The activation of the middle layer, obtained after merging the feature specific layers, gives the universal signature which will be transmitted from the client to the server. Each layer is fully connected with its parents in the network. In our application the output of the network is the facial expressions/attributes to be recognized, one neuron per expression/attribute, with the final values indicating the score for the presence of these attributes.

\subsection{Learning the parameters of the network} 

Carefully setting up the learning of such hybrid network is the main issue for competitive performance. We propose to learn the parameters of this network with a multistage approach. We start by learning an initialization of the common parameters. To do this we work with the most discriminate feature type (e.g. A, B or C). For example, suppose we observed that A  is the most discriminate for our application (as discussed in the experiment section, we will see that for our application FVs are the most discriminant features). 
Thus we start learning the parameters of the network corresponding to both (i) the feature specific parameters of network A (blue layers) and (ii) the part of the network common to all features (black layers).  Then we fix the common parameters and learn the feature specific parameters of the feature B taking training examples encoded with B. In our case, the task is same but the features are different during each training round. By repeating the same procedure, we learn 
the feature specific parameters of the network for each of the remaining type of features. In the end, all the features are aligned into a common signature which can then be transmitted to the server for the computation. 

The major advantage of this strategy is that although we are mapping all the features into same feature space, we do not require feature to feature correspondence e.g.\ we are not using a certain feature type to estimate or mimic any other feature type. Moreover, when we encounter a new feature type, we can easily branch out the existing network and learn its parameter without hindering the performance of other feature types. Thus the proposed learning strategy, while performing very well, also avoids the retraining of the whole network upon addition of a new features type. This is a major 
advantage of this our approach over existing Mod-drop~\cite{neverova2015oddrop} algorithm. Finally, since there are fewer parameters to optimize than training one distinct network per feature, the computations required are less and the training is faster.

Another alternative, that we explored, is to learn the parameters of the whole network first with all the available feature types, and then fix the common parameters 
and fine-tune the feature specific parameters. The reason behind this approach is to make shared subspace more discriminative than with the one learned with the single 
most discriminative feature so that we can align all the component features in this subspace and improve the  overall performance. We found the performance obtained with 
this approach is slightly better than the one we discussed before. However, this alternative requires feature to feature correspondence mapping. Moreover, training with all 
the features at a time requires more computing resource and also leads to slow convergence and longer training time. We compare the performances of these methods 
in more details in the experiment section. 

\begin{figure*}[t]
\centering
\includegraphics[width=0.7\textwidth, trim=0 25 0 0, clip]{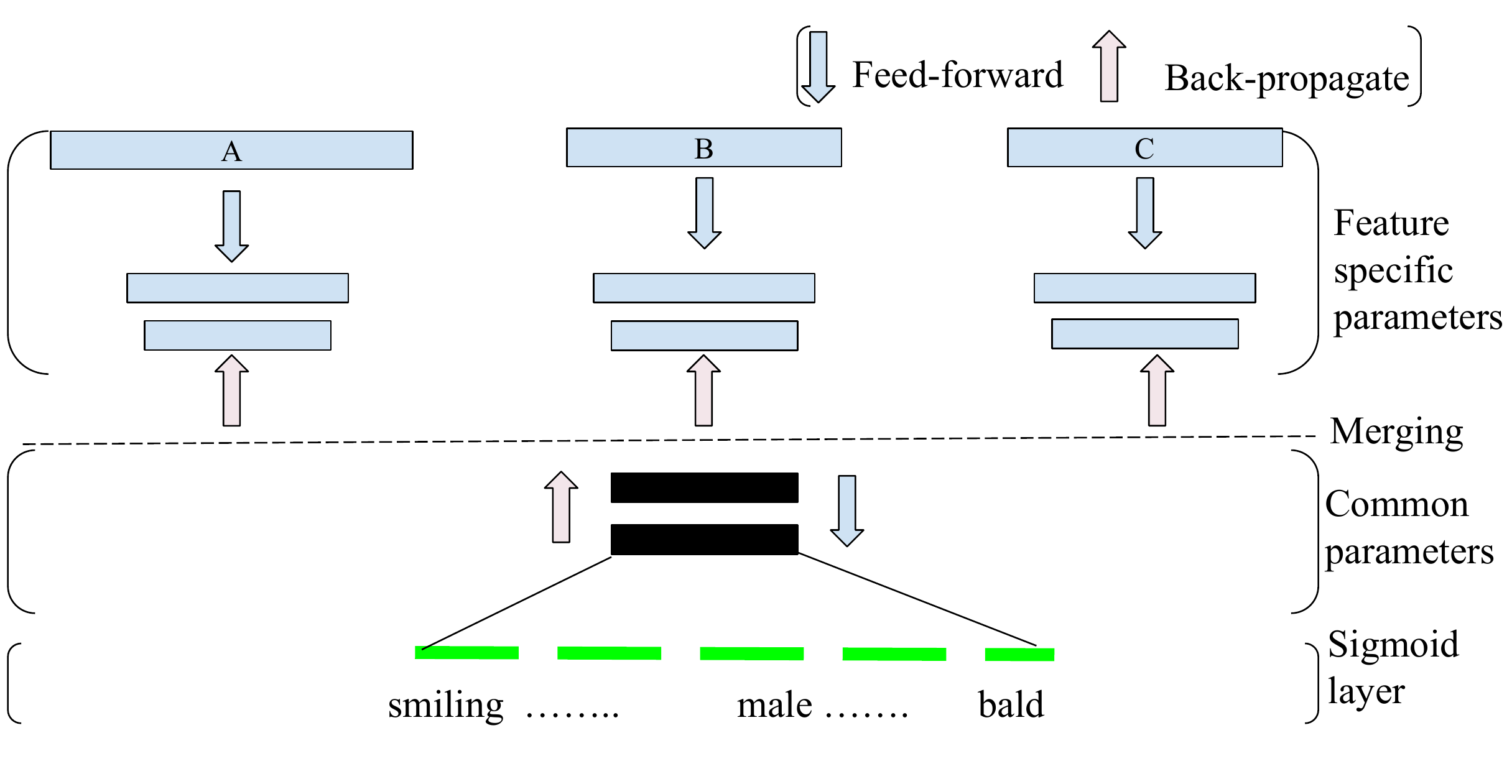} 
\caption{
     Illustration of proposed method. 
 }
\label{fig:illus}
\end{figure*}

%
    
    
    
    
    
    


\subsection{Details of the architecture}
The proposed network is composed of only fully connected (FC) layers. Once the features are fed into the network, they undergo feature specific linear projections followed by processing with Rectified Linear Units (ReLU). Eq.~\ref{eqn:featSpProj} gives the feature-specific transformations, where $\sigma$ is the non-linear 
transformation function i.e.\ ReLU, $W_A, W_B, W_C$ and $\textbf{b}_A, \textbf{b}_B, \textbf{b}_C$ are projection matrices and biases for the input features of the networks A, B, and C respectively. These representations further go into linear projections followed by ReLU depending upon the depth of the network.  
\begin{align}
    & h^A = \sigma(\x_AW_A + \textbf{b}_A )  \nonumber \\
    & h^B = \sigma( \x_BW_B + \textbf{b}_B ) \nonumber \\
    & h^C = \sigma( \x_CW_C + \textbf{b}_C )  
    \label{eqn:featSpProj}
\end{align}

When the network takes more than one type of features at a time, it first transforms them with the FC and ReLU layers and then sums them  and feeds into the common part of the network. We call this step as {\em merging}, as shown in the diagram. We further call the vector obtained at this point, after merging, as the signature of the face.

In the common part of the network, intermediate hidden layers are projected into linear space followed by  ReLU. The final layer of the network is a sigmoid layer. Since we are doing multilabel predictions, sigmoid will assign higher probabilities to the ground truth classes. We learn the parameters to minimize the 
sum of binary cross-entropy of all the predictions of the sigmoid layer. We minimize the loss function using Stochastic Gradient Descent (SGD) with standard back propagation method for network training. 

In the heterogeneous client-server setting, the client is expected to compute the signature and send it to the server for processing. Since different clients can have very different computing capabilities they can compute their signature with different types and number of features -- in the worst case with just one feature. The method allows for such diversity among clients and as the server side works with the provided signature while being agnostic about what and how many features were used to make it.
 
 \begin{table}
\centering
\newcolumntype{K}{>{\centering\arraybackslash}p{6em}}
\newcolumntype{C}{>{\centering\arraybackslash}p{2em}}
\begin{tabular}{|K|c|C|C|C| }
  \hline
  Parameters Type        &   Layer Type  & A      &   B  & C     \\       
  \hline
         &  Input   & $\x_A$      &  $\x_B$  & $\x_C$     \\       
  \multirow{2}{5em}{\centering Feature Specific}  &  FC(ReLU)     	  &	 4096       &   4096     & 4096    \\
                     &  FC(ReLU)             &  1024       & 1024       & 1024  \\
  \hline 
  Merge  & Add & \multicolumn{3}{c|}{1024}   \\

  \hline 
   \multirow{3}{5em}{\centering Common}         &  FC(ReLU)            &  \multicolumn{3}{c|}{1024} \\
    &  FC(ReLU)            &  \multicolumn{3}{c|}{1024} \\
                         &  Sigmoid            &  \multicolumn{3}{c|}{40} \\

\hline 
\end{tabular}
\caption{ Details of parameters of proposed network } 
\label{network_params}
\end{table}

\section{Experiments}
\label{sec:experiments}
We now present the experimental validation of the proposed method on the task of facial attribute classification. All the quantitative evaluation is done on the CelebA dataset~\cite{liu2015deep}, the largest publicly available dataset annotated with facial attributes. There are more than 200,000 face images annotated with 40 facial attributes. This dataset is split into train, val, and test sets. We use train and val set for training and parameter selection respectively, and we report the results obtained on the test set.

In the rest of the section, we first give the implementation details and then discuss the results we obtained.

\subsection{Implementation details}
We have performed all our experiments with 
the publicly available aligned and cropped version of the CelebA\footnote{http://mmlab.ie.cuhk.edu.hk/projects/CelebA.html}~\cite{liu2015deep} dataset (without any further pre-processing). We assume that up to 3 different types of features can be computed, namely, Local Binary Patterns, Fisher Vectors and Convolutional Neural Networks features, as described below. 
\vspace{0.8em} \\
\textbf{Local Binary Patterns (LBP).} We use the publicly available
$\texttt{vlfeat}$~\cite{Vedaldi2008} library to compute the LBP descriptors. The images are cropped to $218\times 178$ pixels. We set cell size equal to $20$, which yields a descriptor of dimension 4640.
\vspace{0.8em} \\
\textbf{Fisher Vectors (FV).} 
We compute Fisher Vectors following Simoyan et al~\cite{simonyan2013fisher}. We compute dense SIFTs at multiple scales, and compress them to a dimension of 64 using Principal Component Analysis. We use a Gaussian mixture model with 256 Gaussian components. Thus, the dimension of the FV feature is of 32,768 (2$\times$256$\times$64). The performance of this descriptor is $77.6 \pm 1.2\%$ on LFW for the task of face verification, with unsupervised setting, which is comparable to the one reported~\cite{simonyan2013fisher}. 
\vspace{0.8em} \\
\textbf{Convolutional Neural Networks (CNN).}
We use the publicly available state-of-art CNN mode trained on millions of faces presented in~\cite{parkhi2015deep}, to compute the CNN features. The dimension of CNN feature is of 4096. Our implementation of this feature gives $94.5 \pm 1.1\%$ on LFW for verification in unsupervised setting. Here, these features are computed without flipping and/or multiples of cropping of faces.

\subsection{Baseline methods.} 
We report two different types of baselines. In the first one, the network is trained with a given feature type (e.g. LBP) while the same type of feature is used at test time (e.g. LBP again). We call this type of network as \emph{Dedicated Networks}. In the second setting, we allow the set of features at train time and the one used at test time to differ. Such networks are adapted to different sets of features. This is the particular situation we are interested in. More precisely, we experimented with 3 different dedicated networks (one per feature type) and 2 adapted networks, as detailed below, all such are considered as baselines. 
\vspace{0.8em} \\
\textbf{LBPNet/FVNet/CNNNet.}
These baseline networks use only LBP, FV or CNN features, respectively, for both training and testing. They provide the single feature performances, assuming that no other feature is available either at training or testing.
\vspace{0.8em} \\
\textbf{All Feature Training Network (AllFeatNet).}
In this setting, all the available features are used to train the network. At test time, one or more than one type of features can be used, depending on its availability. For us, the available features are as described before FVs, CNNs, and LBPs. 
\vspace{0.8em} \\
\textbf{Mod-Drop.} This is currently the best method for learning cross-modal architectures, inspired by \cite{neverova2015oddrop}. It consists, at train time, in randomly sampling a batch of examples including only one type of features at a time, instead of taking all the available features, and learn the parameters in a stochastic manner. We refer the reader to the original work \cite{neverova2015oddrop} for more details.
 
\subsection{The proposed method.}
\begin{figure}[t]
\centering
\includegraphics[width=0.9\columnwidth, trim=0 5 0 0, clip]{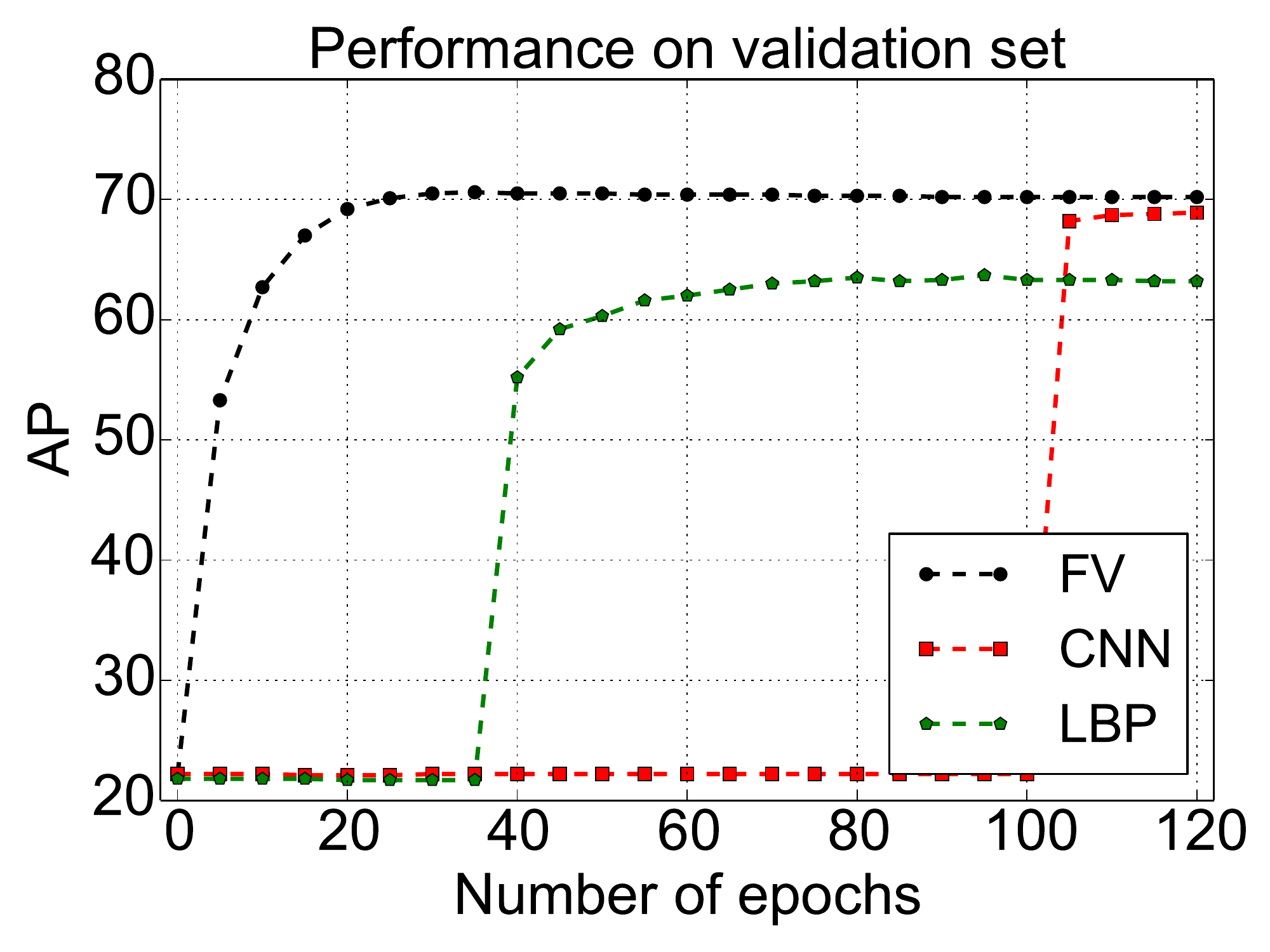}
\vspace{-1em}
\caption{
  Performance of FVs, CNNs, and LBPsmeasured on the validation set.
}
\label{fig:optm_curve}
\end{figure}

On the basis of which we fix the parameters of the common shared subspace, we categorize the proposed methods into two:
\vspace{0.8em} \\
\textbf{FVNetInit.} Tab.~\ref{tab:perf_single_feat} shows the individual performance of different features we used for our experiments. From the table we can see that FVs are most discriminative for our application. Thus, we choose to take few top layer's parameters ( please refer Tab.~\ref{network_params} of for the number of layers in shared subspace ) of FVNet as common shared parameters of proposed network. Once we fix this, we learn the feature specific parameters for CNNs and LBPs to minimize the loss function. Fig.~\ref{fig:optm_curve} shows the evolution of performances of FVs, LBPs, and CNNs with the amount of training epochs. 
%
\vspace{0.8em} \\
\textbf{AllFeatNetInit.} In this case, we use the common part of AllFeatNet as a starting point. Then we fix these parameters and learn the feature specific parameters of FVs, LBPs and CNNs to minimize the loss the function. 
\vspace{0.8em} \\
\subsection{Quantitative results}
We now present the results of the experiments we do to evaluate the proposed method. We measure the performance using average precision (AP) i.e.\ the area under the precision vs.\ recall curve. 
We do not consider attribute label imbalances for all the cases, unless explicitly stated. 

\begin{table}[tb]
\centering
\newcolumntype{K}{>{\centering\arraybackslash}p{10em}}
\newcolumntype{C}{>{\centering\arraybackslash}p{10em}}
\begin{tabular}{|K|C| }
  \hline
  Method       & Avg. Precision\\
  \hline
  Random       & 23.1\%       		  \\
 
  FVNet     & \textbf{69.0\%}     \\
 
  CNNNet    &  68.7\%             \\
  
  LBPNet    & 64.3\%             \\
    
  \hline
\end{tabular}
\caption{Average Precision (AP) of single feature type baselines} 
\label{tab:perf_single_feat}
\end{table}

\begin{table}[tb]
\centering
\newcolumntype{K}{>{\centering\arraybackslash}p{10em}}
\newcolumntype{C}{>{\centering\arraybackslash}p{10em}}
\begin{tabular}{|K|C| }
   \hline
  Method       & mean Avg. Precision\\
  \hline
  AllFeatNet             &   63.4 $\pm$ 9.5 \% 		  \\
 
  Mod-Drop            &   67.8 $\pm$ 3.7 \%    \\
 Ours(FVNetInit)       &   68.8 $\pm$ 3.0\%        \\
 
  Ours(AllFeatNetInit)    &  \textbf{69.0 $\pm$ 3.4\%}      \\
  \hline
\end{tabular}
\caption{mean AP(mAP) of  multi-feature baselines} 
\label{tab:perf_multi_feat}
\end{table}

\begin{table*}[tb]
\centering
\newcolumntype{K}{>{\centering\arraybackslash}p{4.5em}}
\newcolumntype{C}{>{\centering\arraybackslash}p{8em}}
\begin{tabular}{|K|C|C|C|C|C| }
  \hline
  Features       & Dedicated Network & AllFeatNet        &   Mod-Drop                &  Ours (FVNetInit) & Ours \hspace{3em} (AllFeatNetInit) \\
  \hline
  FV          &  69.0\%        	  &	 64.2\% (-4.7\%)	 &   70.0\% (+1\%)   &  68.7\% (-0.3\%) &  68.8\% (-0.2\%)  \\
 
  CNN         & 68.7\%       	  &  63.3\% (-5.5\%)  &   68.2\% (-0.5\%) &  68.1\% (-0.6\%) & 67.9\% (-0.8\%)  \\
 
  LBP         &  64.3\%          &  42.5\% (-21.8\%) &  59.6\% (-4.7\%)  & 62.1\% (-2.2\%) & 61.5\% (-2.8\%)     \\

  \hline 
\end{tabular}
\caption{ Comparing the proposed methods with other methods using dedicated networks. The table shows that the performance of the proposed methods is competitive to the one of dedicated networks, while the performance of other compared methods is significantly low, particularly in the case of LBPs.  } 
\label{tab:compare_with_dedicated_network}
\end{table*}

\begin{figure}[tb]
\centering
\includegraphics[width=0.9\columnwidth, trim=0 0 0 0, clip]{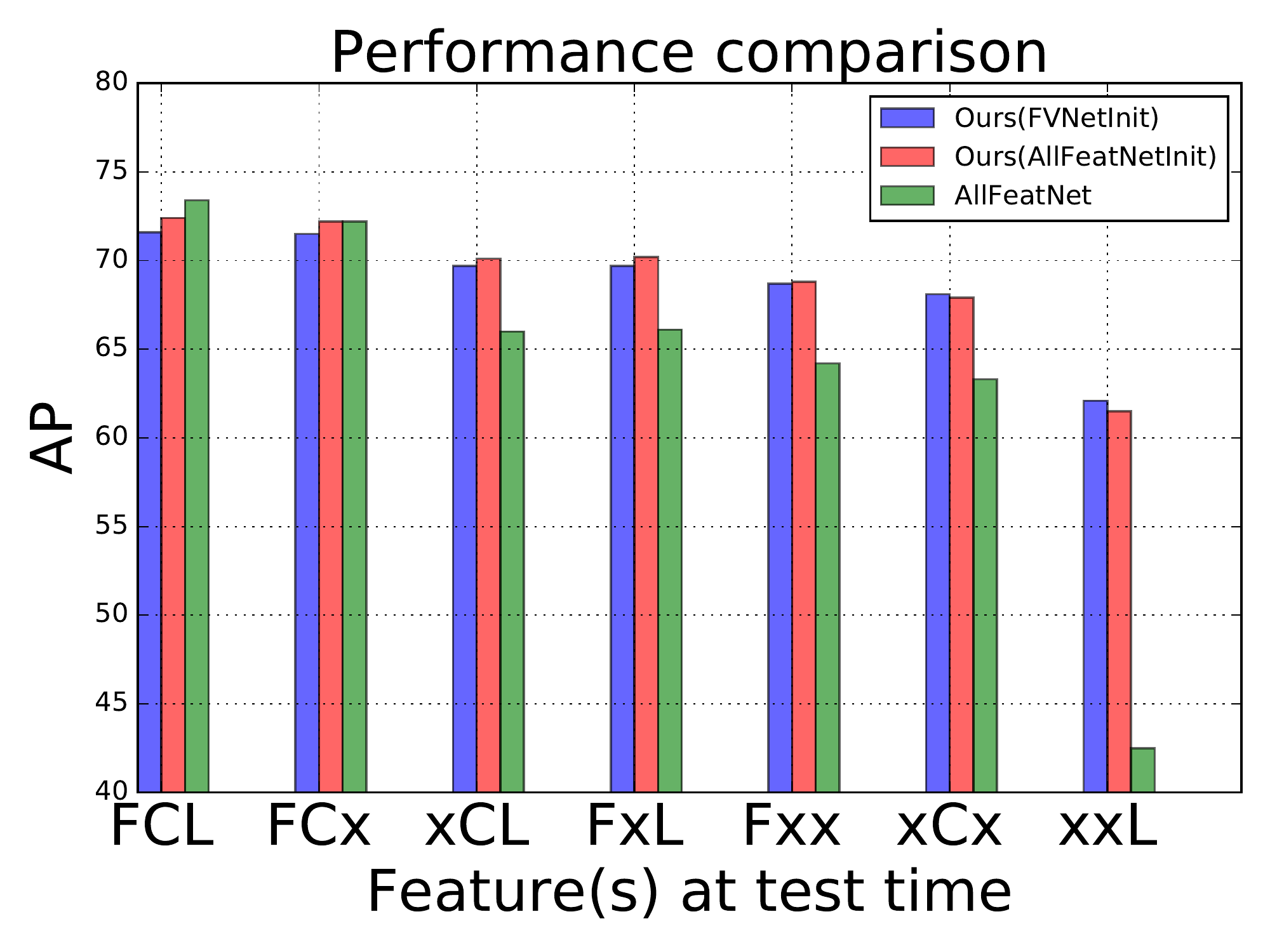} 
\vspace{-1em}
\caption{
  Performance comparison between different methods and different combinations of feature(s) at test time. FCL represents FVs, CNNs, and LBPs respectively. \textbf{'x'} denotes the absence of the  corresponding feature.
}
\label{fig:perf_compare}
\end{figure}

Our experiments are mainly focused on validating two aspects of the proposed method. First, we demonstrate that the performance due to individual features are retained after merging all the features in the same common subspace. Second, we demonstrate that the performance is improved in the presence of more information, i.e.\ presence of multiple types of features at a time.
\vspace{0.8em} \\
\textbf{Performance comparison with Dedicated Networks.}
Tab.~\ref{tab:perf_single_feat} and Tab.~\ref{tab:compare_with_dedicated_network} give the performance of single features networks and  their comparison with that of the multi-feature trained network (when, at  test time, only one type of feature is present). From these tables, we can observe that, with both our approaches, the performance of the component features at test time is competitive to that of dedicated networks trained with those features only.  
Compared to existing methods such as Mod-Drop and AllFeatNet, the range of  performance drops in comparison to dedicated networks is the least in our case. More precisely, the widest drop range for us is up to $-2.8\%$ w.r.t.\ that of LBPNet in AllFeatNetInit network. While for the same feature, it is up to $-4.7\%$ in Mod-Drop and up to $-21.8\%$ in AllFeatNet w.r.t.\ that of LBPNet. These results clearly 
demonstrate that our method is more robust in retaining the performances of individual features while projecting them in common subspace.
%
\vspace{0.8em} \\
\textbf{Performance comparison with Multi-feature Networks.}
Table~\ref{tab:perf_multi_feat} compares the mean average precision (mAP) of different multiple features based networks with the proposed method. For a network with $3$ different types of input features, there are $7$ different possible combinations of feature(s) at test time. The performance shown in the table is the mean AP obtained with all these combinations. The proposed method outperforms the other multi-feature-based networks. This shows that the proposed network and the multi-stage training strategy is capable of making better predictions in the presence of more information i.e.\ multiple types of features at a time and are optimal to every combination of features. 

Fig.~\ref{fig:perf_compare} shows the performance comparison between the proposed methods with AllFeatNet at different levels of feature combinations. From the bar-chart, we can observe that, when all the features are available at  test time, AllFeatNet performs better than ours. It is expected too, because this approach is optimized only for this combination. But this is the most unlikely scenario for the applications we are addressing, due to constraints such as computing resources and time, etc. Out of other $6$ cases, our method performs substantially better and gives similar performance in one case. This shows that our method leverages all the features available and when more information is present, gives better performance. Unlike AllFeatNet, the proposed method is optimal in every combination of features too.

\subsection{Qualitative results}
\begin{figure*}[t]
\centering
\includegraphics[width=1.0\textwidth, trim=0 -5 0 10, clip]{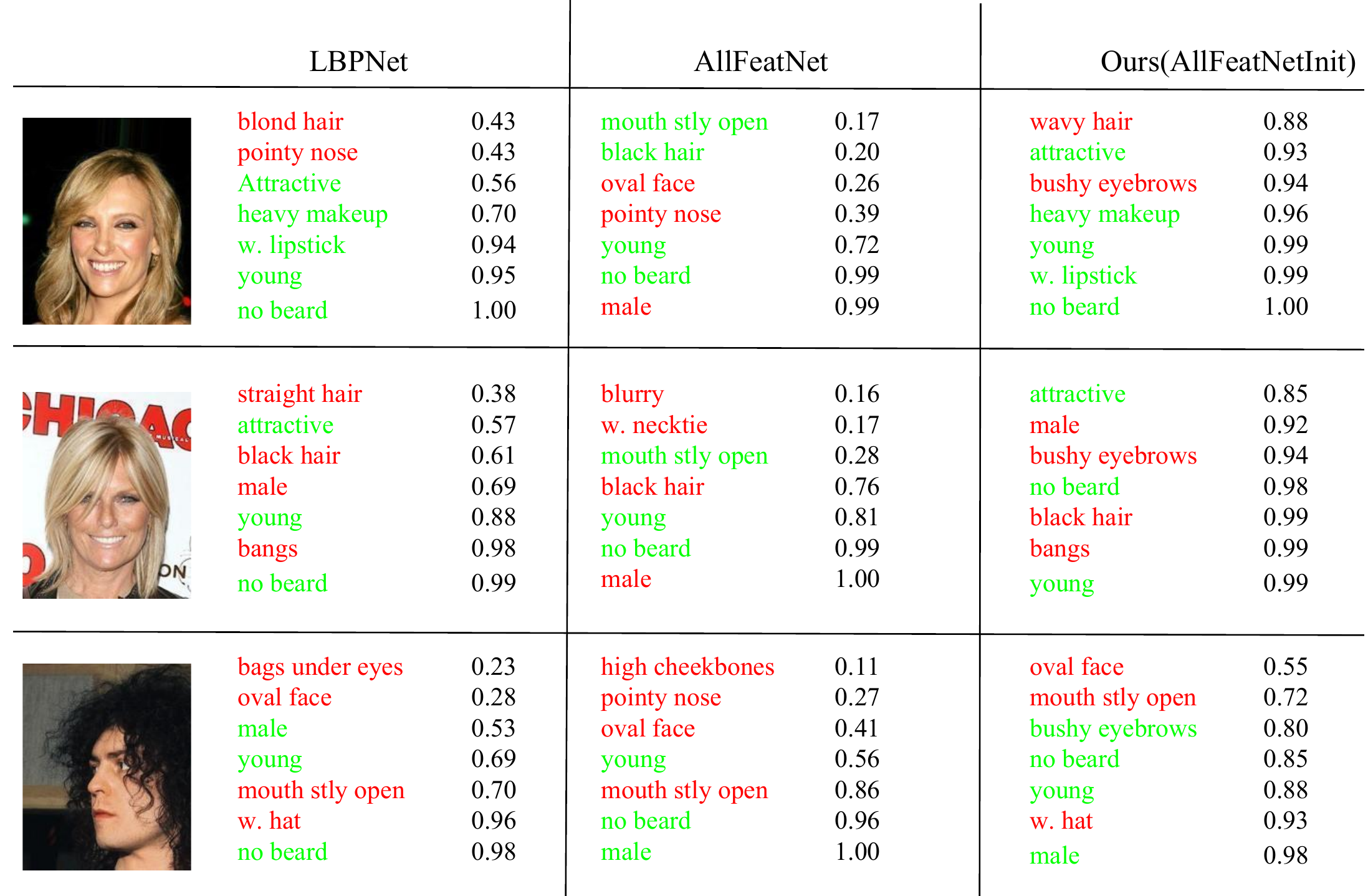} 
\caption{
  Qualitative results comparison of the proposed method with other methods. Top 7 attributes predicted by these methods are shown. As before green color indicates relevant attributes whereas red color indicates irrelevant attributes for the image. (Better viewed in color)
}
\label{fig:qualit_eval}
\end{figure*}
Fig.~\ref{fig:qualit_eval} shows the qualitative performances comparison between the baselines and the proposed method. We randomly choose three different test images and used them for evaluation. 
Here, we consider LBPs (the simplest feature type) only for evaluation. Thus for both the single feature network (LBPNet) and multi-feature network ( AllFeatNet and ours), only LBPs are available at test time. In the figure we can see the top $7$ attributes 
predicted by the compared methods. For each of the attributes, the corresponding score shows the probability of an attribute being present in the given image.
On the basis of the number of correct predicted attributes, the performances of LBPNet and the proposed method is comparable in two cases (first two cases). While in the third case, our method ($4$ correct predictions ) is even better than LBPNet ($3$ correct predictions). This further validates that the proposed method retains the property of component features.
 The performance of AllFeatNet is comparatively poorer than LBPNet and ours for all test images. Moreover, it is important to note that the scores corresponding to the predicted attributes by AllFeatNet  are small. This suggests that with this approach the predictive power of LBPs is masked by other strong features e.g. FV and CNNs. 
 
\section{Conclusions}
\label{sec:concls}
We propose a novel hybrid deep neural network and a multistage training strategy, for facial attribute classification. We demonstrated, with extensive experiments, that the proposed method retains the performance of each of the component features while aligning and merging  all the features in the same subspace. In addition to it, when more than one feature type is present, it improves the performance and attains state-of-art performance. The proposed method is also easily adaptable to new features simply learning the feature specific parameters. This avoids retraining the existing network. Since the majority part of the network is shared among all the feature types, the proposed method reduces the number of parameters.

\vspace{.5em}{\bf Acknowledgments} This project is funded in part by the ANR (grant ANR-12-SECU-0005). 
\bibliographystyle{splncs}
\bibliography{biblio}
\end{document}